# Microscopic Vehicle Trajectory Datasets from UAV-collected Video for Heterogeneous, Area-Based Urban Traffic

Yawar Ali [a], K. Ramachandra Rao [b*], Ashish Bhaskar [c], Niladri Chatterjee [d]

[a] *Transportation Research & Injury Prevention Centre (TRIPC), Indian Institute of Technology Delhi, New Delhi-110016, India, Tel.: +91 9582684383 Email: aliyawar274@gmail.com*

[b] *Department of Civil and Environmental Engineering, Indian Institute of Technology Delhi, New Delhi-110016, India, Tel.: +91-11-2659-1235 Email: rrkalaga@civil.iitd.ac.in*

[c] *Faculty of Engineering, School of Civil & Environmental Engineering, Queensland University of Technology, Brisbane, Australia, Tel.: +61 73138 9985 Email: ashish.bhaskar@qut.edu.au*

[d] *Department of Mathematics, Indian Institute of Technology Delhi, New Delhi-110016, India, Tel.: +91-11-2659-1490 Email: niladri@maths.iitd.ac.in*

## Abstract

This paper offers openly available microscopic vehicle trajectory (MVT) datasets collected using unmanned aerial vehicles (UAVs) in heterogeneous, area-based urban traffic conditions. Traditional roadside video collection often fails in dense mixed traffic due to occlusion, limited viewing angles, and irregular vehicle movements. UAV-based recording provides a top-down perspective that reduces these issues and captures rich spatial and temporal dynamics. The datasets described here were extracted using the Data from Sky (DFS) platform and validated against manual counts, space mean speeds, and probe trajectories in earlier work. Each dataset contains time-stamped vehicle positions, speeds, longitudinal and lateral accelerations, and vehicle classifications at a resolution of 30 frames per second. Data were collected at six mid-block locations in the national capital region of India, covering diverse traffic compositions and density levels. Exploratory analyses highlight key behavioural patterns, including lane-keeping preferences, speed distributions, and lateral manoeuvres typical of heterogeneous and area-based traffic settings. These datasets are intended as a resource for the global research community to support simulation modelling, safety assessment, and behavioural studies under area-based traffic conditions. By making these empirical datasets openly available, this work offers researchers a unique opportunity to develop, test, and validate models that more accurately represent complex urban traffic environments.

# 1. Introduction

The collection of detailed traffic data is essential for transportation research, including areas such as traffic engineering, travel behaviour analysis, and simulation-based modelling. Traditionally, collecting detailed traffic data has been a demanding and resource-intensive task, constrained by challenges related to accuracy, cost, and scalability. However, image processing and artificial intelligence (AI) have revolutionised data extraction methods, offering highly automated, scalable, and precise solutions. The ability to extract macroscopic and microscopic traffic variables from aerial imagery has opened new possibilities in traffic monitoring, congestion assessment, and various transportation system analyses.

Reliable traffic safety assessment depends critically on the quality and granularity of trajectory data. For conflict-based and perception-aware surrogate safety analysis, microscopic vehicle trajectory (MVT) data, which includes high-fidelity time-stamped vehicle positions, serves as the fundamental input. In heterogeneous, area-based traffic environments, collecting such data is challenging due to the disregard of lane discipline, diverse vehicle geometries, and frequent lateral movement patterns. This paper aims to provide high-quality MVT data from multiple urban midblock locations for the researchers to use.

## 1.1 Fixed-Camera Methods and Manual or Semi-Automated Data Extraction

Traditionally, video data has been collected using fixed surveillance cameras, mounted on poles or nearby infrastructure, that overlook road sections or intersections. These systems are most effective in lane-based environments where occlusion is minimal and vehicle movement is predictable. Fixed roadside CCTV cameras have been essential for traffic monitoring. The type of information collected and the methods used to collect it differ significantly based on the objectives (Kuciemba and Swindler, 2016). Researchers often focus on gathering crucial traffic metrics, such as vehicle counts and movement tracking (Zangenehpour et al., 2015). These data are employed not only to analyse traffic flows but also to conduct safety studies concerning risky interactions at various traffic facilities (Hu et al., 2004; Morris et al., 2012; St-Aubin et al., 2015, 2013). Researchers have focused on detecting and tracking moving objects, particularly as technology has rapidly advanced (Buch et al., 2011; Tian et al., 2011).

Manual extraction of trajectories, although laborious, was common in early research studies. Analysts would annotate vehicle positions across successive frames, calculate speeds and gaps, and derive basic safety metrics such as Time to Collision (TTC) or Post-Encroachment Time (PET) (Archer, 2005). While highly accurate in low-volume scenarios, manual methods are:

- Time-consuming, often taking days to extract a few minutes of usable data.
- Prone to human error, particularly under occlusion or dense traffic.
- Infeasible at scale, especially for empirical model calibration.

## 1.2 Emergence of Semi-Automated Tools for Data Extraction

With the introduction of advanced computer vision and trained models, semi-automated tools such as Traffic Data Extractor (TDE) and Data from Sky (DFS), these tools typically:

- Track each vehicle path with mouse clicks and save the output to a CSV.

- Use frame differencing or object tracking algorithms.
- Require manual correction with smoothing techniques for missed or misclassified objects.
- Offers good accuracy in ordered traffic but degrades in an area-based heterogeneous environment.

Studies demonstrated how such systems could be associated with surrogate safety software (e.g., SSAM) to produce large datasets of conflict events (Archer, 2005; Gettman et al., 2003). However, performance deteriorates when applied to area-based, heterogeneous traffic, where lateral movement, unexpected accelerations, and occlusions are frequent. Several studies (Kadali and Vedagiri, 2013; Kanagaraj et al., 2015) observed that fixed-camera views often result in:

- Occlusion of smaller vehicles like motorcycles or rickshaws behind larger vehicles.
- Perspective distortion, particularly at the edges of the frame.
- Field-of-view limitations, restricting the usable area for interaction analysis.

These issues prompted the exploration of aerial data collection and automated extraction methods.

**1.3 UAV-Based Data Collection and Automated Extraction**

Unmanned Aerial Vehicles (UAVs) enable high-resolution aerial surveillance, supporting real-time traffic monitoring, vehicle tracking, and congestion detection. UAVs offer a unique vantage point for collecting traffic data, particularly in environments where traditional fixed cameras or roadside sensors are insufficient or ineffective. Over the past decade, researchers have utilised UAV-based data collection to enhance traffic flow modelling, road safety assessments, and dynamic traffic management strategies (Lee and Kwak, 2014). Integrating computer vision with AI-based data extraction enables automated classification, trajectory tracking, and vehicle speed estimation, among other variables.

Over the past two decades, extensive research has been conducted on vehicle detection and classification using video data. Two primary methods are employed from the field of image processing to accomplish this task (Buch et al., 2011). The top-down (TD) approach and the bottom-up (BU) approach. The TD approach involves identifying vehicle geometry based on its motion characteristics and classifying it into predefined categories through set rules. Key techniques for isolating the foreground include frame differencing, background subtraction, Gaussian mixture models (GMMs), and graph cuts, with background subtraction being preferred due to its effectiveness (Huang et al., 2017). Machine-learning techniques classify these foreground elements, including artificial neural networks (ANNs), support vector machines (SVMs), and nearest-neighbour classifiers. These methods typically train on motion features, such as corners, edge maps, and optical flow (MacKay, 2012). The BU approach varies significantly from the TD approach as it leverages specific object features and detects alterations in pixel values, subsequently categorising these changes as parts of an object. These components are then aggregated into complete objects for vehicle detection. Techniques for identifying interest points, which are pixel locations where local features are gathered, include the basic path method, scale-invariant feature transform (SIFT), and histogram of oriented gradients (HOG). These interest points are subsequently classified and assimilated into objects using machine learning techniques such as artificial neural networks (ANNs), support vector machines (SVMs), and boosting methods. TD methods have some limitations when it comes to urban traffic, due to challenges such as misclassification caused by shadows and under- (or over-) classification of vehicles due to occlusion by larger vehicles on smaller following vehicles. BU algorithms are increasingly preferred in such traffic states because they are more adept at managing issues like shadows and occlusions (Li et al., 2013).

Recent developments have incorporated cloud computing technologies into the design of systems that facilitate the management and analysis of traffic monitoring data (Abdullah et al., 2014). Systems capable of autonomously processing and analysing video streams through a GPU cluster have been proposed. These systems utilise a cascade classifier for vehicle detection (Abdullah et al., 2014). The techniques employed for vehicle tracking include Kalman filtering, particle filtering, spatial-temporal Markov random fields, and graph correspondence (Coifman et al., 1998; Morse et al., 2016). Analysing driver behaviour from video data entails converting the microscopic trajectories (MVTs) of individual vehicles into comprehensive descriptions of their behaviour and interactions with other vehicles (Interacting vehicle pairs) that primarily rely on detecting vehicles and tracking their movements (Morris and Trivedi, 2013).

Recent advancements in neural network-based image recognition techniques have resulted in significant improvements in detection accuracy (Krizhevsky et al., 2012; Simonyan and Zisserman, 2015). These methods are adept at identifying various types of objects across different scales. Traditional techniques segmented an input image into a fixed grid, where each segment was analysed for a single object label and subsequently merged with adjacent segments (Ciregan et al., 2012; Szegedy et al., 2013). However, contemporary methods utilise neural networks not only to detect objects but also to predict their precise locations, resulting in more accurate and comprehensive outcomes (Redmon et al., 2016; Ren et al., 2015). Deep learning-based methods have recently been introduced for video analytics within innovative city applications (Wang and Sng, 2015). One of the libraries gaining popularity for image detection is YOLO (Redmon et al., 2016). This library stands out for its capability to detect objects quickly, making it suitable for real-time traffic operations and safety applications. YOLO integrates object detection and recognition into a single neural network model, allowing the algorithm to effectively consider the overall information of a frame while being less affected by shadowing effects (Huang et al., 2017).

Although image processing techniques are generally developed independently of video data types, bird's-eye view imagery captured by UAVs has demonstrated significant advantages in terms of the quality of the output produced (Ali et al., 2024a). Earlier studies shed light on how drones have transformed the approach to collecting and analysing visual data. It highlights the versatility and efficiency of UAVs in various applications, particularly in transportation research. The focus on drones underscores a significant shift towards adopting more advanced, aerial-based methods for video analysis, reflecting a broader trend in utilising technology to enhance data accuracy and operational efficiency in multiple aspects. Several researchers have compiled a summary of the most recent global research trends concerning the utilisation of unmanned aerial vehicles (UAVs), commonly known as drones, for traffic monitoring and analysis (Kanistras et al., 2015; Renard et al., 2022; Salvo et al., 2014; Valavanis and Vachtsevanos, 2015). These studies can be divided into two categories based on two different criteria: (i) equipment type and (ii) video processing method. Initially, most traffic-related applications involved either mounted video cameras or fixed-wing quadcopters, whereas in recent years, only a few researchers have initiated investigations using simple multi-rotor UAVs (quadcopters). Initially, most traffic-related applications involved fixed-wing UAVs (Barmpounakis et al., 2016; Khan et al., 2017; Renard et al., 2022). Similarly, two main categories of studies have been developed based on the video-processing approach: (i) Semi-Automatic or manual methods and (ii) Automatic methods. Semi-automatic methods involve precise but laborious and time-consuming processes, as each study requires manual object detection and tracking for many frames (Salvo et al., 2014). However, recently, there has been a significant shift towards using automatic methods in studies, resulting in rapid data processing and evolution. This shift has culminated in the capture of real-time traffic data analysis by UAVs, marking a notable advancement in traffic monitoring and management technologies (Apeltauer et al., 2015; Khan et al., 2017; Renard et al., 2022; Zheng et al., 2015).

Unmanned Aerial Vehicles (UAVs) provide a top-down, unobstructed view of traffic scenes, significantly improving the quality and completeness of trajectory data. Their mobility allows for coverage of wide spatial regions, such as intersections, signalised junctions, midblock locations, and informal roundabouts that are otherwise difficult to monitor. The UAV collected data in this study helped capture dense, heterogeneous traffic streams in Indian cities. This aerial data enables:

- Detection of lateral seepage, side-by-side movement, and irregular lane changes.
- Observation of multi-vehicle interactions over extended fields of view.
- Calibration of perception-based and self-organised models with high temporal and spatial resolution.

A study demonstrated the efficacy of UAVs in tracking multi-class vehicle trajectories with significantly lower occlusion rates than pole-mounted cameras (Khan et al., 2017). However, UAV data requires advanced processing algorithms to translate raw footage into usable MVT datasets.

**1.4 Automated Extraction using Data from Sky**

Data from Sky (DFS) is one of the advanced AI image processing tools that extracts detailed traffic data from video footage. DFS claims a 98 to 100 per cent accuracy range in structured, homogeneous, lane-based traffic environments. While its performance has been validated in well-organised urban traffic systems, its effectiveness in heterogeneous, area-based traffic conditions, such as those prevalent in India and other developing countries, remains largely unexplored. Unlike ordered traffic, heterogeneous traffic is characterised by mixed vehicle types, frequent lateral movements, and area-based interaction dynamics, making accurate data extraction a significant challenge.

An earlier study critically evaluates the applicability and accuracy of DFS in heterogeneous, area-based traffic conditions (Ali et al., 2024a). A comprehensive validation approach is employed, integrating Classified Volume Count (CVC), Space Mean Speeds (SMS) across vehicle categories, and microscopic trajectory tracking using GPS-based probes. A comparative error analysis is performed to assess the discrepancies between DFS-extracted and manually recorded data, with Mean Absolute Percentage Error (MAPE) calculations providing a quantitative measure of accuracy. The findings offer empirical insights into the strengths and limitations of DFS in complex traffic environments, contributing to the ongoing discourse on AI-driven traffic data collection and extraction methodologies.

By bridging empirical evaluation with methodological advancements, the reliability of the Data from Sky tool for traffic data extraction in heterogeneous and area-based traffic settings has been validated (Ali et al., 2024a). The MVT datasets presented in this paper have broader implications for traffic simulation, travel behaviour modelling, and data-driven traffic engineering solutions, offering practical insights for global transportation research.

The process of UAV-based data collection and validation is described in detail in the earlier work (Ali et al., 2024a). The validation of extracted data using DFS is also elaborated in the same article. Overall, DFS achieved reliable accuracy when applied to UAV-collected video of heterogeneous area-based traffic from a bird's eye perspective. This paper focuses on describing the datasets and their structure.

## 2. Data Description and Exploratory Data Analysis

The quality and contextual relevance of MVT datasets directly influence the credibility of safety analysis, behaviour modelling, and simulation accuracy. This section presents the various MVT datasets from heterogeneous and area-based traffic environments that are offered to researchers globally. The datasets are provided as CSV files, each containing frame-by-frame microscopic vehicle trajectory (MVT) data.

Variables include:

- Vehicle ID (unique for each vehicle)
- Timestamp (30 fps)
- Local coordinates in metres (X, Y)
- Length and Width of vehicle in meters (m)
- Speed in km/h
- Longitudinal and lateral acceleration in m/s$^2$
- Vehicle Type (Car, Motorised two-wheeler (M2W), Motorised three-wheeler (M3W), Light commercial vehicle (LCV), Bus, Heavy Commercial Vehicle (HCV))

All datasets are extracted for each frame in a 30-fps video, resulting in 30 data points per second. Table 1 provides their descriptive statistics and discusses key patterns revealed through exploratory data analysis (EDA).

Table 1 Dataset Summary

| Dataset | Location | Duration (min) | Composition (%) | Stretch (m) | Width (m) | Location Code |
|---|---|---|---|---|---|---|
| Delhi (UAV Collected) | Midblock | 63 | 41:34:15:8:1:1 | 180 | 14 | DEL_2 |
| Delhi (UAV Collected) | Midblock | 63 | 40:31:15:11:2:1 | 180 | 14 | DEL_3 |
| Delhi (UAV Collected) | Midblock | 66 | 41:40:11:5:1:1 | 200 | 10.5 | DEL_4 |
| Delhi (UAV Collected) | Midblock | 66 | 35:45:11:6:1:2 | 200 | 10.5 | DEL_5 |
| Noida (UAV Collected) | Midblock | 41 | 41:39:10:4:1:5 | 170 | 14 | NDA_1 |
| Noida (UAV Collected) | Midblock | 41 | 38:39:11:5:1:6 | 170 | 14 | NDA_2 |

*Composition: Car:M2W:M3W:LCV:Bus:HCV*

The datasets capture variability in terms of speed distributions, vehicle fleet composition, flow, density, and, importantly, varied driver behaviours. The offered datasets are extracted using the DFS software from video data collected through UAVs, providing a bird's-eye perspective. Each dataset is discussed in terms of its location, traffic conditions, vehicle composition, and duration.

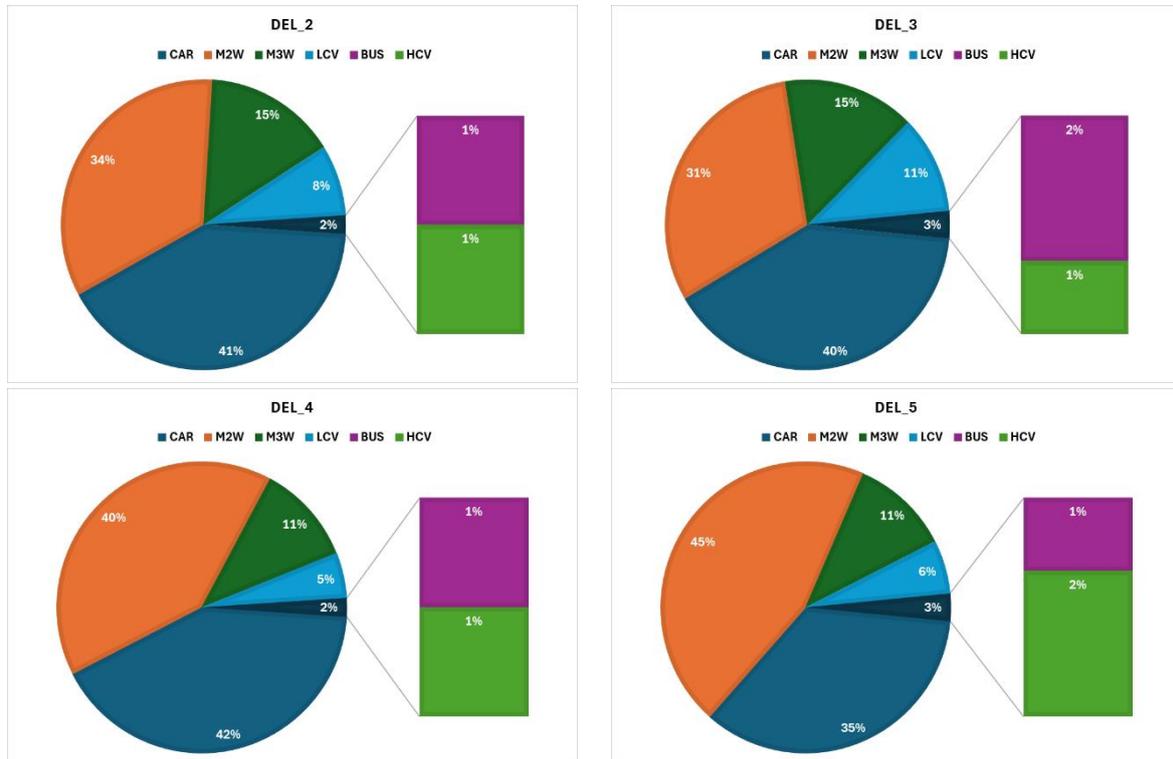

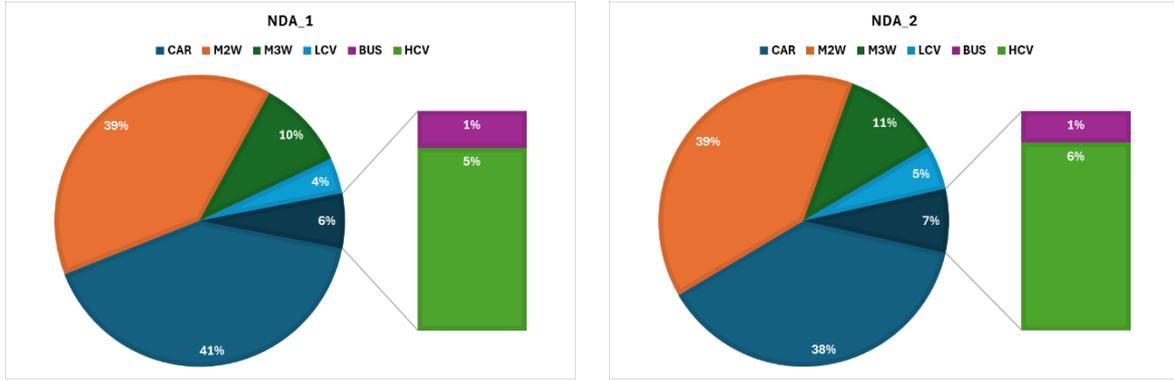

Figure 1 Vehicle Types Distribution

## 2.1 Descriptive Statistics

Table 2 presents descriptive statistics for vehicle speed and acceleration across a diverse set of locations, for UAV-collected empirical datasets. The figures provide a comparative overview of traffic conditions in various urban locations, revealing considerable variation in both speed and acceleration metrics.

Table 2 Descriptive Statistics of the Datasets

| Location Code | Statistics | Speed (km/h) | Acceleration (m/s2) |
|---|---|---|---|
| DEL_2 | Mean | 33.53 | 0.059 |
| | Median | 33.92 | 0.051 |
| | S.D. | 6.92 | 0.299 |
| | Range | 38.31 | 1.677 |
| DEL_3 | Mean | 11.22 | -0.016 |
| | Median | 9.72 | -0.006 |
| | S.D. | 7.57 | 0.436 |
| | Range | 33.30 | 2.339 |
| DEL_4 | Mean | 34.93 | -0.0032 |
| | Median | 35.09 | 0.0013 |
| | S.D. | 8.43 | 0.3299 |
| | Range | 46.88 | 1.822 |
| DEL_5 | Mean | 36.08 | 0.1298 |
| | Median | 36.06 | 0.1273 |
| | S.D. | 9.64 | 0.2839 |
| | Range | 53.33 | 1.5776 |
| NDA_1 | Mean | 33.27 | -0.0007 |
| | Median | 33.93 | 0.0062 |
| | S.D. | 18.41 | 0.4088 |
| | Range | 88.48 | 2.2374 |
| NDA_2 | Mean | 28.09 | 0.0607 |
| | Median | 30.09 | 0.0303 |
| | S.D. | 17.10 | 0.4099 |
| | Range | 77.36 | 2.2146 |

One of Delhi's datasets (DEL_3) exhibits very low average speeds (11.22 km/h) with minimal variation in acceleration, consistent with heavily congested traffic environments.

In contrast, DEL_2, DEL_4, and DEL_5 record significantly higher average speeds (33.5 to 36.1 km/h) and a more stable acceleration pattern. The consistency between median and mean values in DEL_5 suggests stable traffic conditions.

Noida datasets (NDA_1 and NDA_2) display high speed variability and relatively low average acceleration. The high standard deviation in speed indicates inconsistent movement, resulting from intermittent merging and U-turning traffic.

The table provides a comparative insight into traffic characteristics across different datasets. Urban areas tend to exhibit slower, more variable traffic with frequent deceleration. The variation in acceleration also highlights the diverse traffic dynamics, various vehicle types, and the differing levels of infrastructure service.

## 2.2 Data visualisation

The datasets were explored using descriptive statistics and visualisations to highlight key features of heterogeneous urban traffic. The following plots illustrate speed, acceleration, and lane-keeping behaviours across sites and vehicle types. These examples demonstrate the richness of the data and provide initial insights into how the datasets may be used in further research.

### 2.2.1 Speed distribution plots

The speed distribution at different locations used in this study is shown in the following figures. The distribution of speeds shows a different nature at each selected location. The speed distributions reveal a distinct pattern among sites. DEL_3 shows a strong concentration at low speeds, consistent with a congestion state. DEL_2, DEL_4, and DEL_5 present bell-shaped distributions centred around 30-35 km/h, indicating smoother flows. NDA_1 and NDA_2 exhibit flatter distributions with long tails, indicating that vehicles experience both free-flow states and slowdowns.

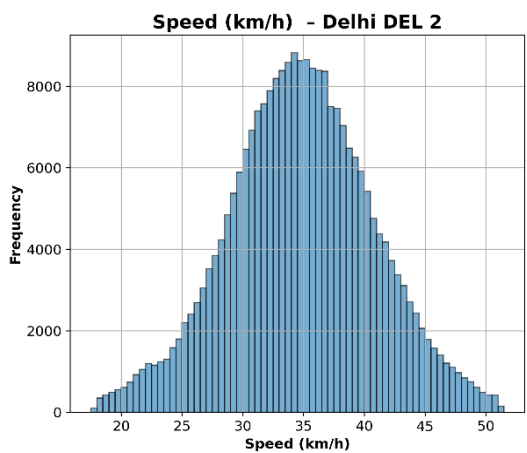
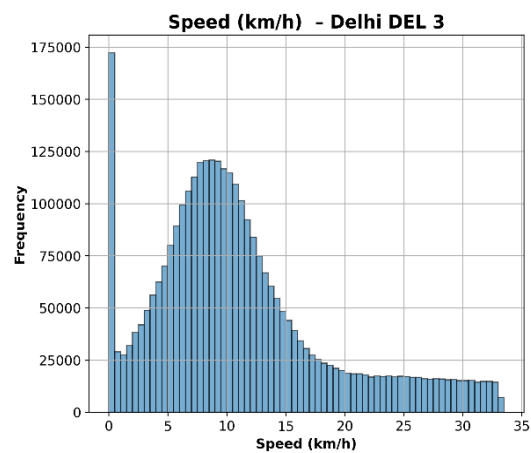

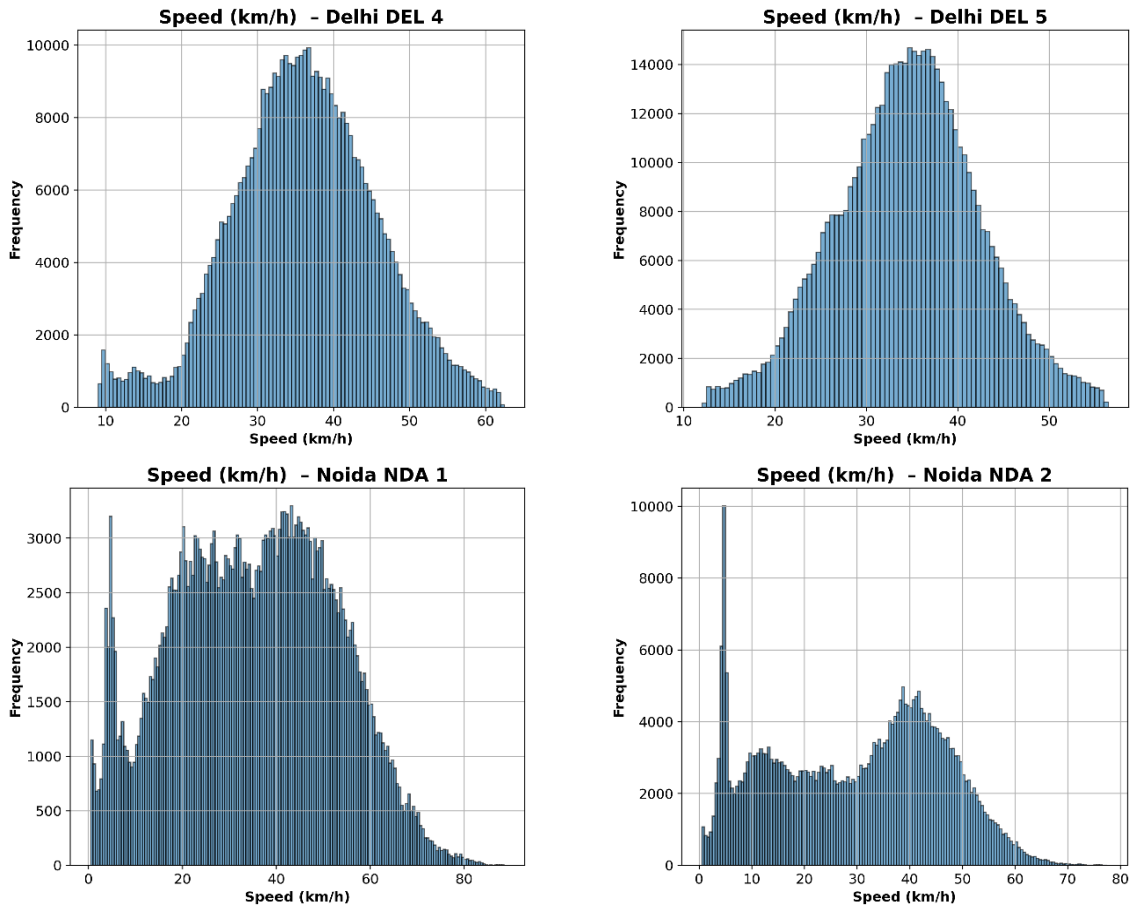

Figure 2 Speed distribution plots

### 2.2.2 Acceleration distribution plots

The longitudinal acceleration distribution plots are shown for all the locations. The distribution for all locations peaks around zero (0), indicating that the majority of vehicles tend to maintain their speeds. This pattern is in concurrence with the on-site behaviour of the population; everyone aims to drive "normally" unless necessary to evade a conflict. This observation also supports the extreme value phenomenon. Extreme events are rare and are concentrated around the tail of the distribution. The rare but significant values in the tails of the distributions are especially important for safety studies, since they capture conflict situations and evasive actions.

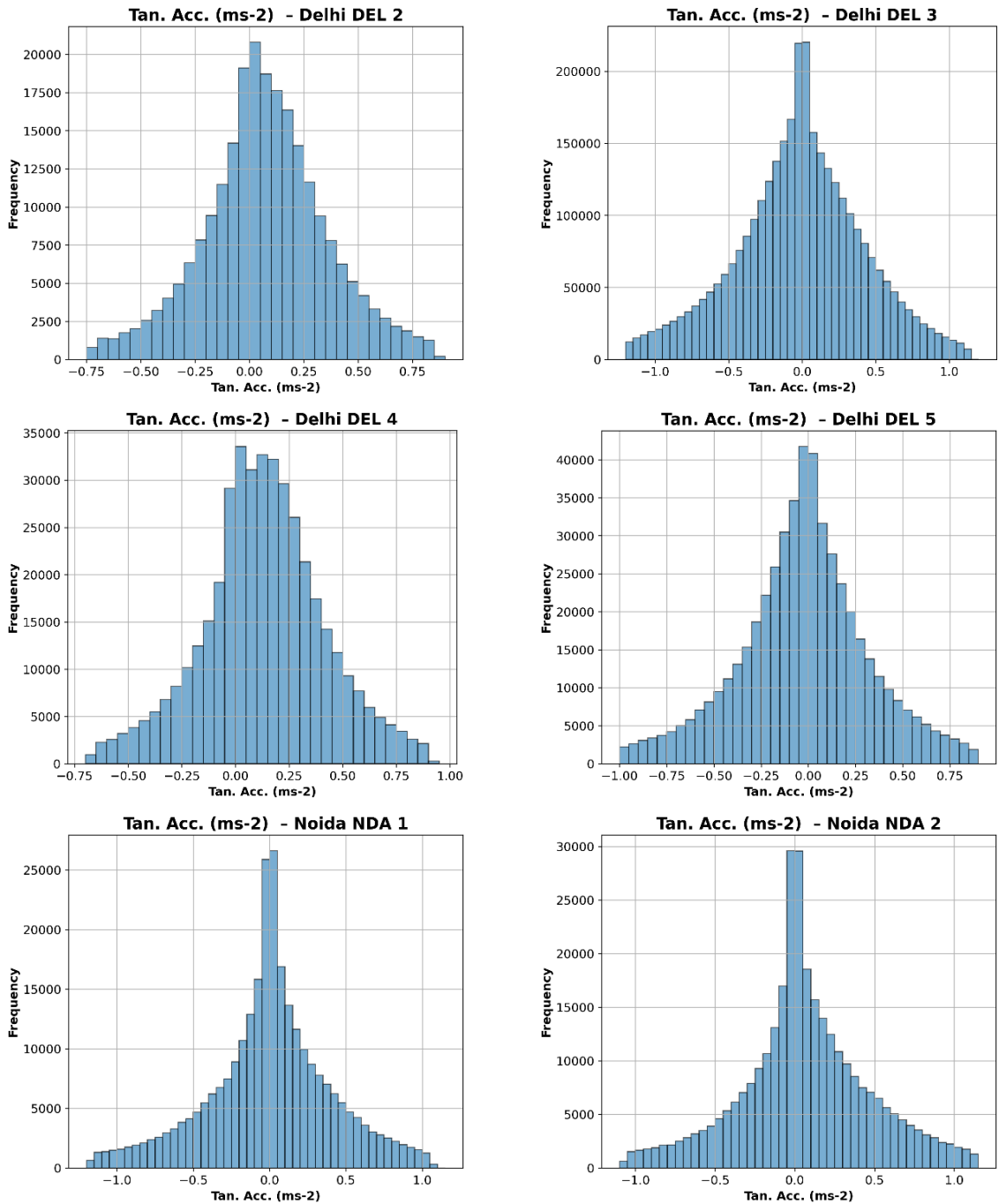

Figure 3 Longitudinal acceleration distribution plots

### 2.2.3 Speed distribution by vehicle types

The following heat plots show the distribution of speeds by vehicle types. The majority of the fleet, comprising cars, M2Ws, and MThWs, shows that most of the heat is distributed among them. The heat plots reveal that two-wheelers (M2Ws) consistently occupy the higher-speed range compared to cars and three-wheelers (M3Ws). Cars form the majority of the mid-speed distribution, while heavier vehicles such as buses and HCVs remain at the lower end. This distinction by vehicle type reflects the agility of two-wheelers in finding space within the stream, compared to the stability of larger vehicles.

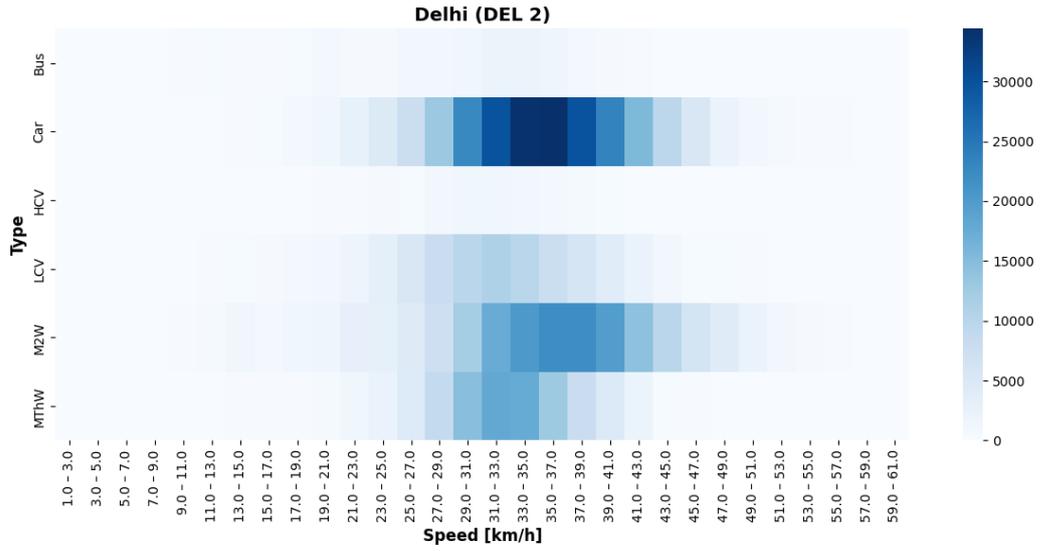
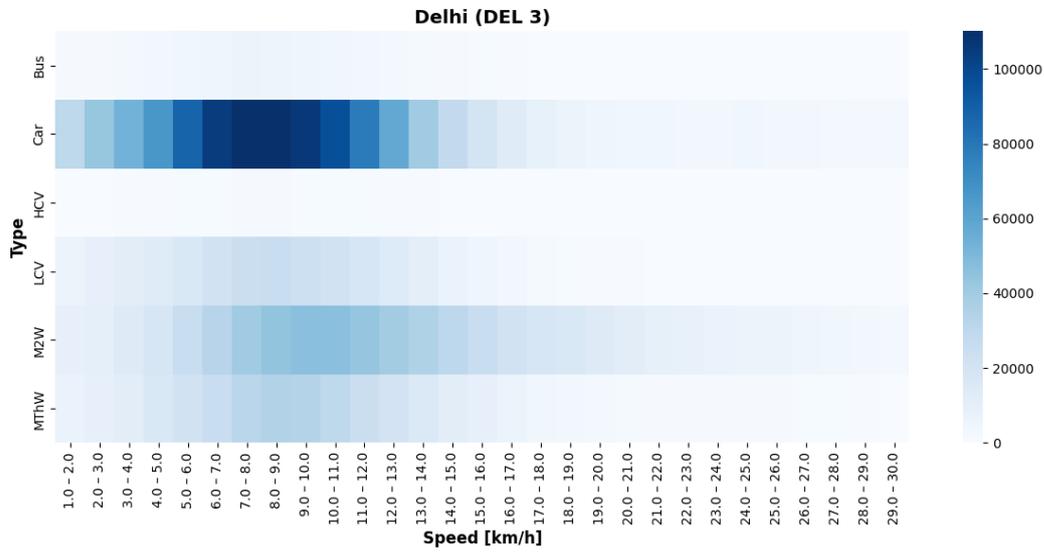
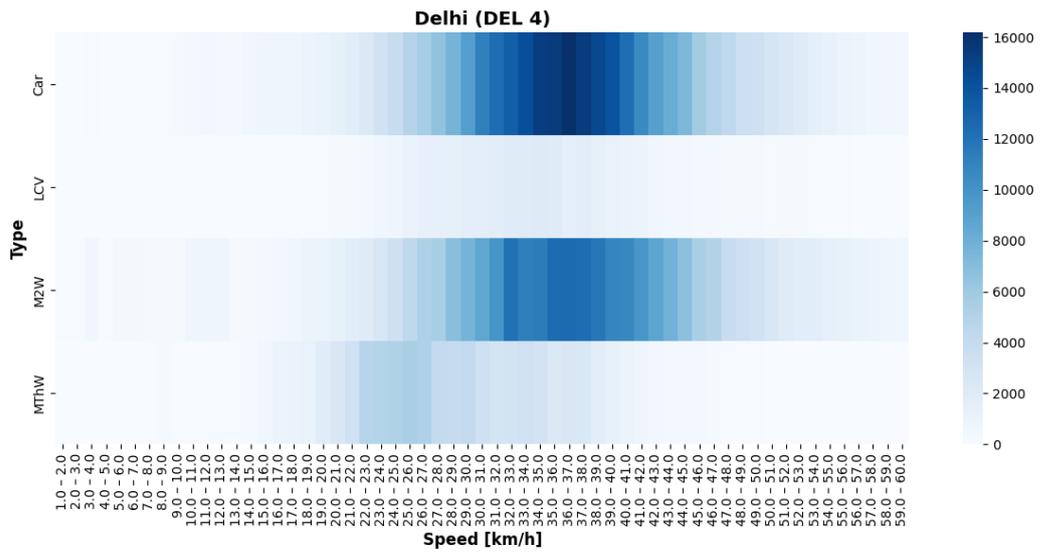

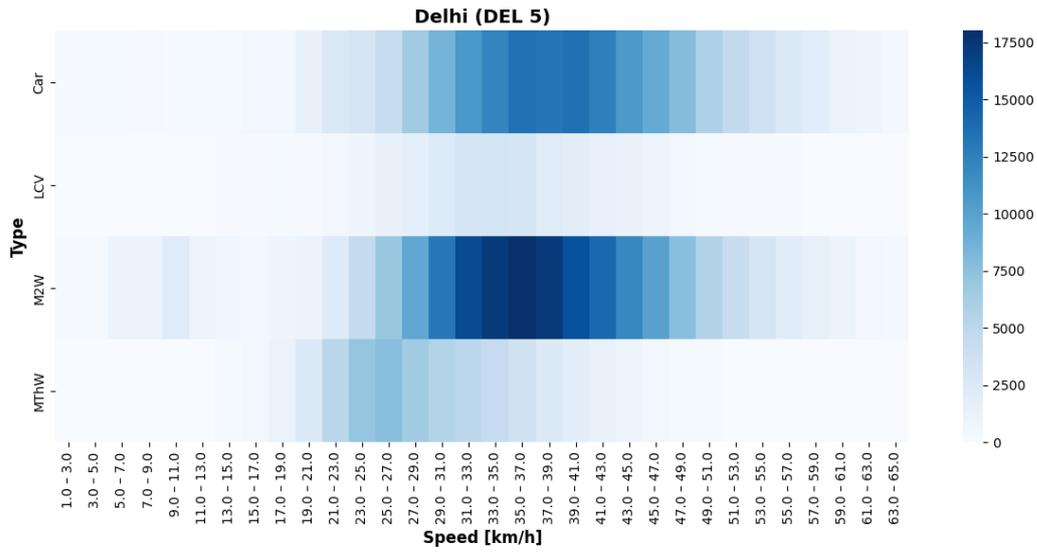
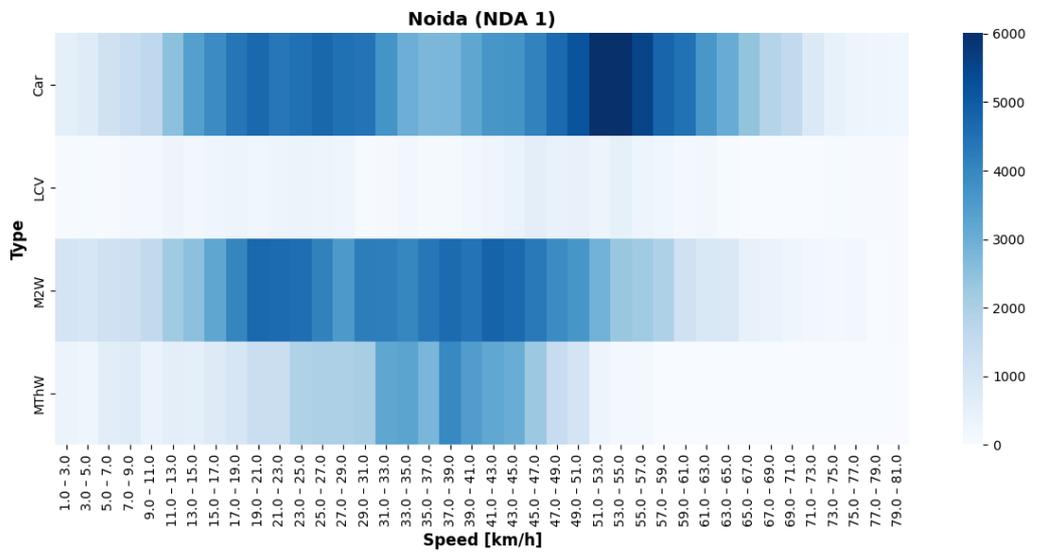
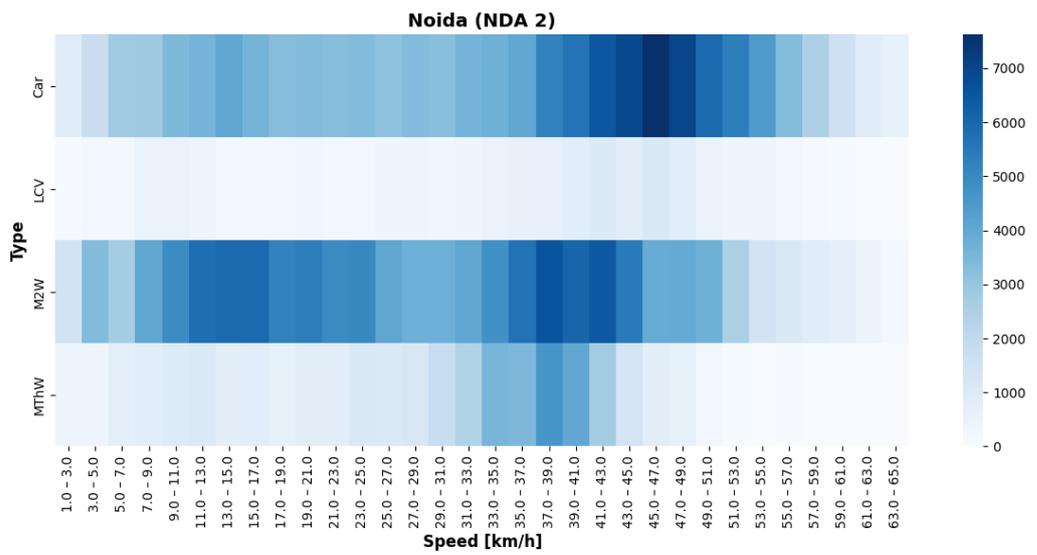

Figure 4 Speed distribution by vehicle types

## 2.2.4 Acceleration distribution by vehicle types

The following plots show the distribution of longitudinal acceleration by vehicle types. The most common observation is that M2Ws have the highest frequency of positive longitudinal acceleration amongst all, demonstrating agile and aggressive driving behaviour. Cars and heavier vehicles show a more concentrated distribution, reflecting smoother but slower acceleration patterns. This distinction makes the datasets valuable for modelling class-specific driving behaviour in heterogeneous area-based traffic.

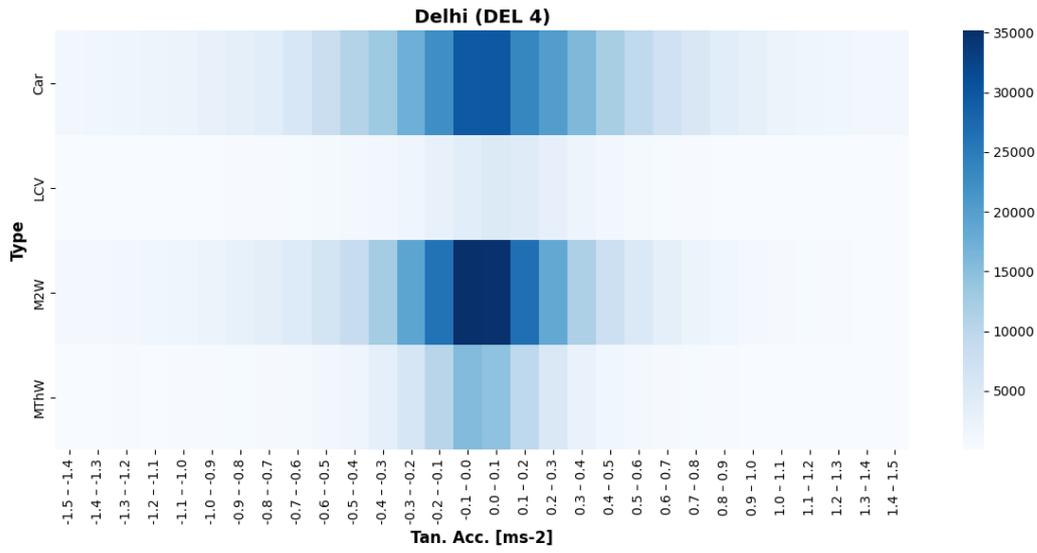
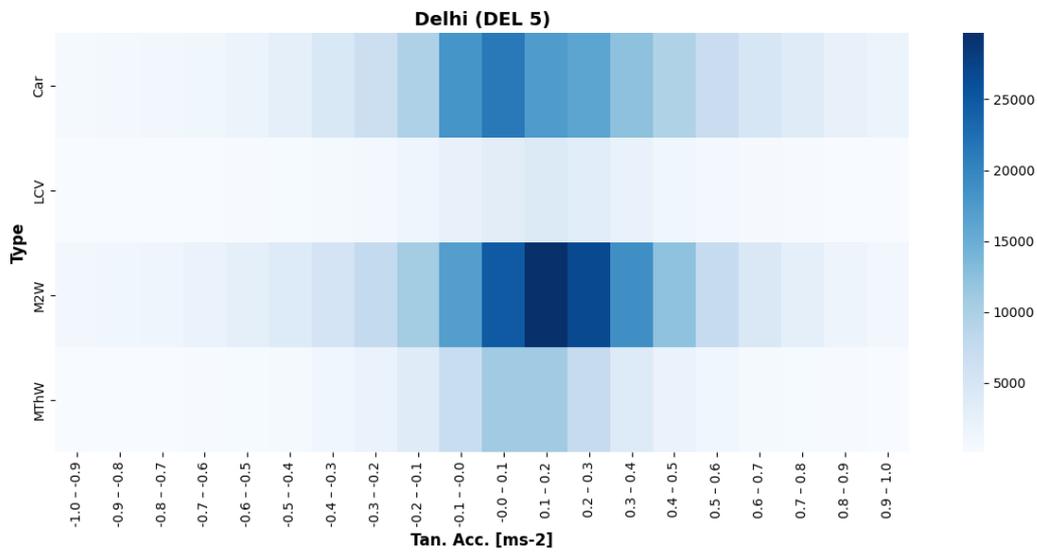
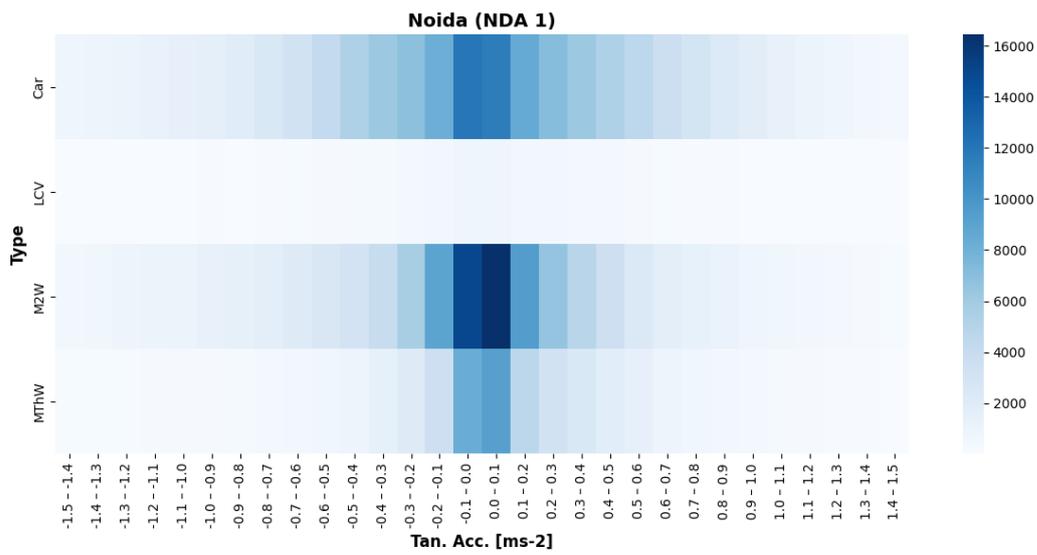

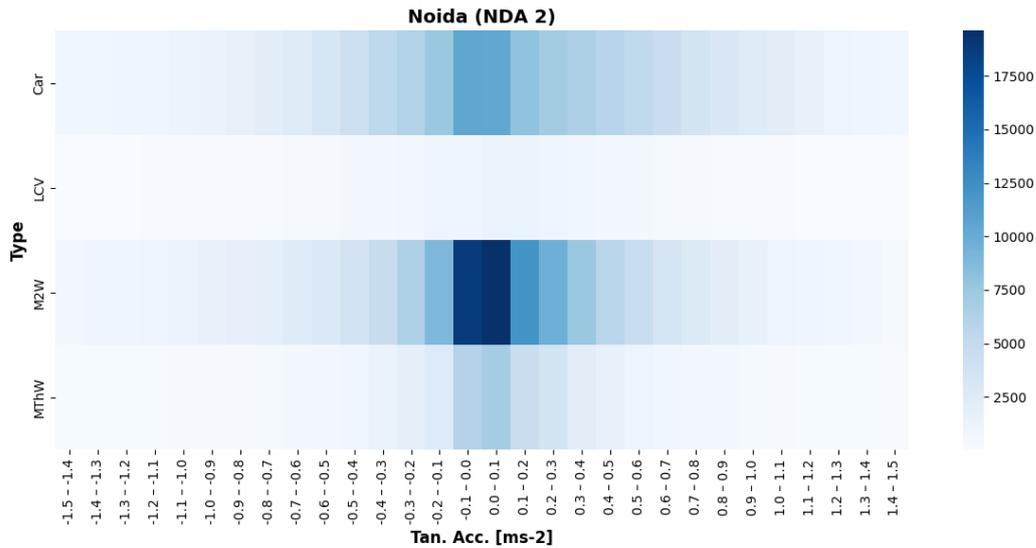

Figure 5 Longitudinal acceleration distribution by vehicle types

### 2.2.5 Lateral acceleration distribution by vehicle types

The following plots show the distribution of lateral acceleration by vehicle types. The lateral acceleration distributions clearly separate vehicle types. Cars cluster tightly around zero, reflecting more disciplined lateral movement. In contrast, two-wheelers show a wide distribution, including significant values away from zero. This is consistent with weaving and filtering behaviour often observed in dense, heterogeneous area-based traffic. Such lateral dynamics are critical for realistic traffic simulation and safety assessment.

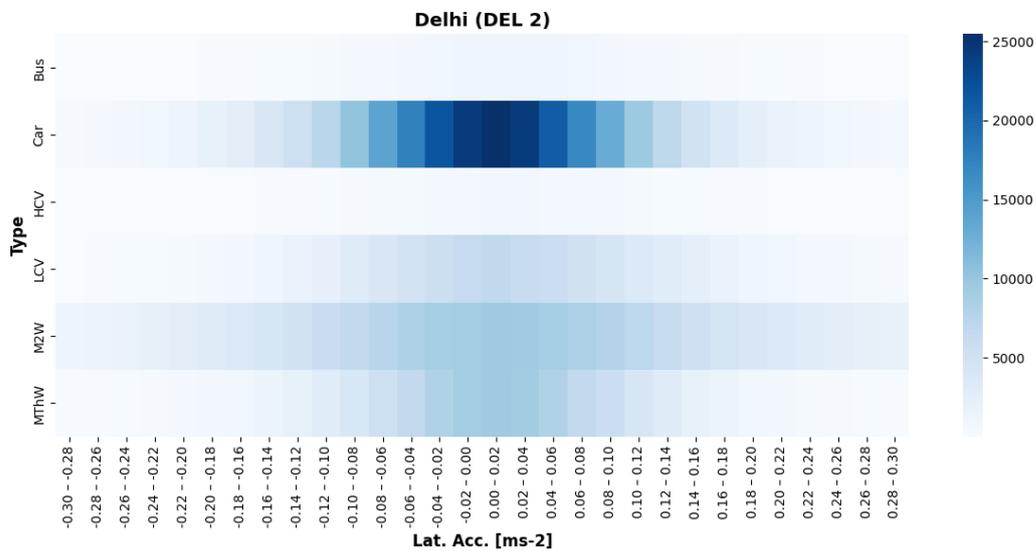

## Delhi (DEL 3)

## Delhi (DEL 4)

## Delhi (DEL 5)

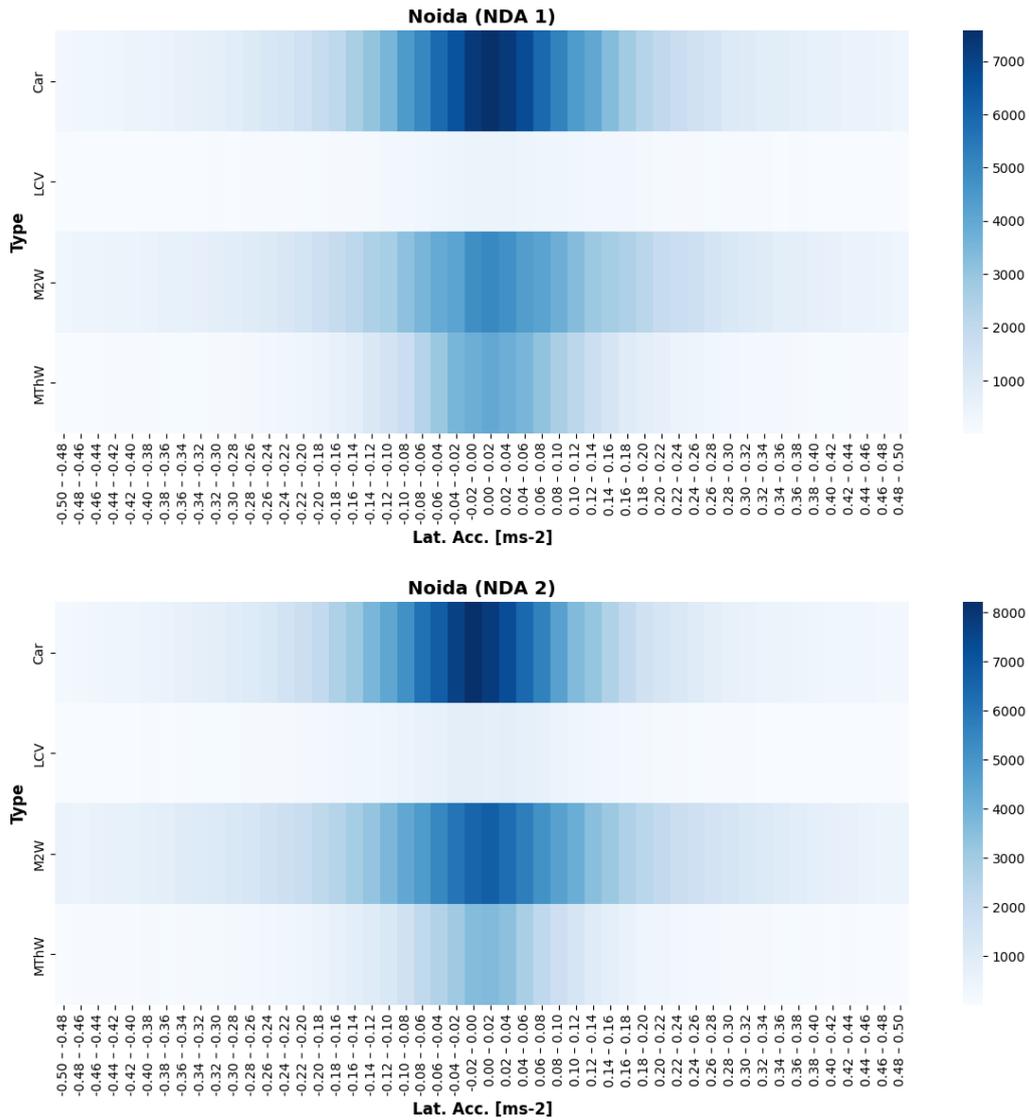

Figure 6 Lateral acceleration distribution by vehicle types

### 2.2.6 Lane keeping trends

The following plots visualise the lane keeping trends across the road width by different vehicle types. The 0 value is the median side speed lane; the higher the value of the x-axis, the closer the vehicle is to the kerb side lane. The most common observation, similar to what is observed in the literature (Ali et al., 2024b; Kanagaraj et al., 2015), is that cars predominantly prefer to stay on the median side, the speed lane. While M2Ws are more comfortable in keeping to the kerb side lanes. These spatial preferences highlight self-organisation patterns in heterogeneous traffic, where each vehicle type adapts differently to the same infrastructure.

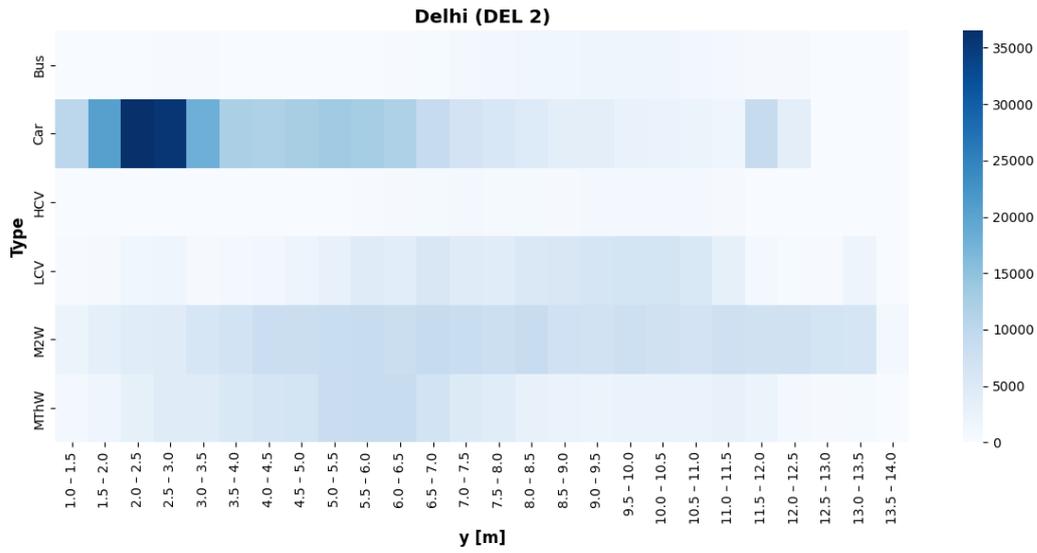
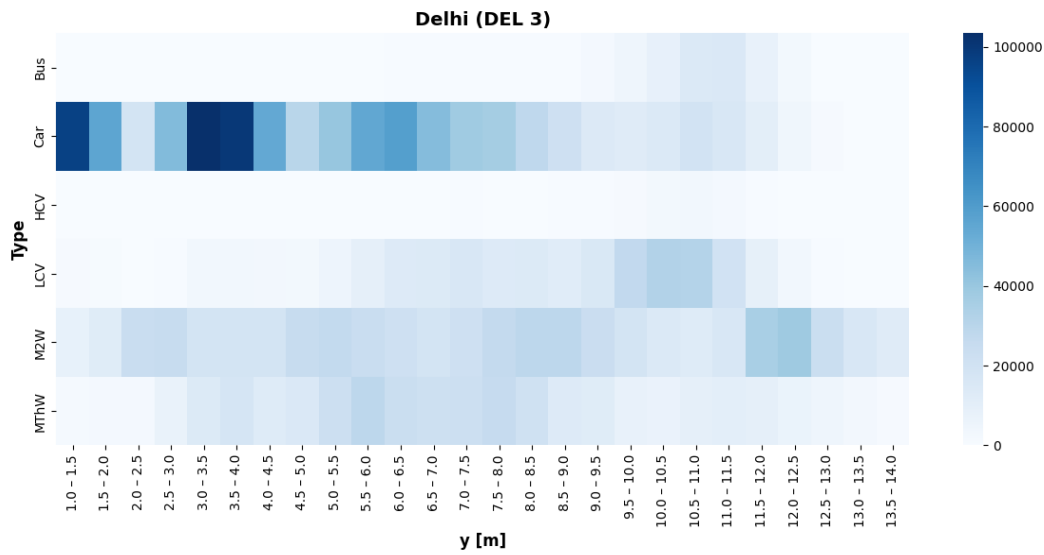
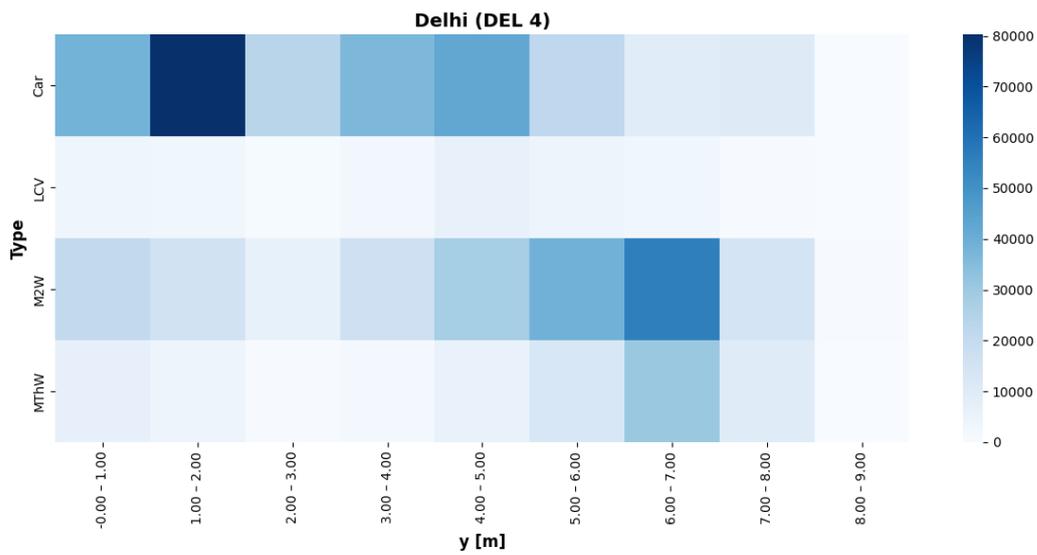

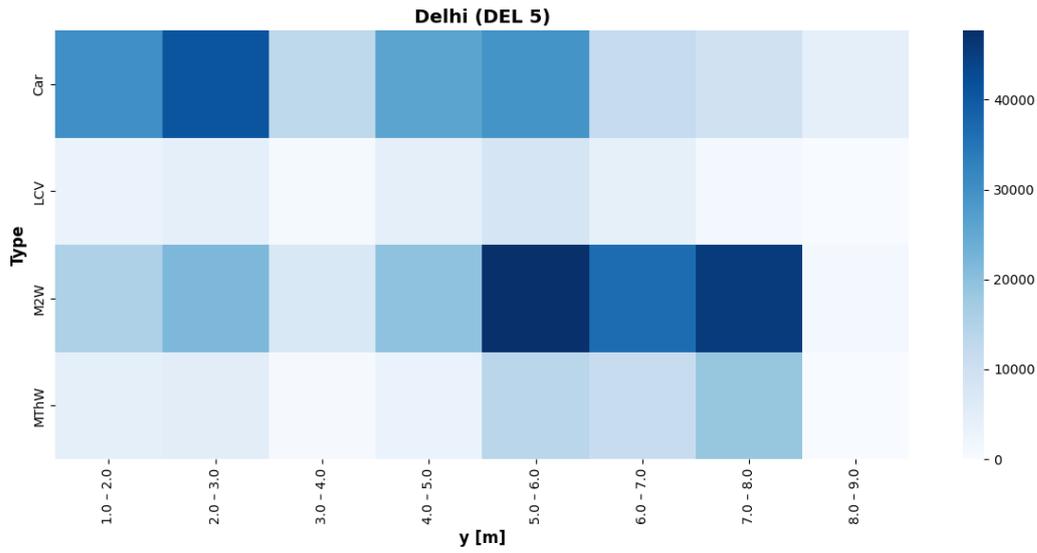
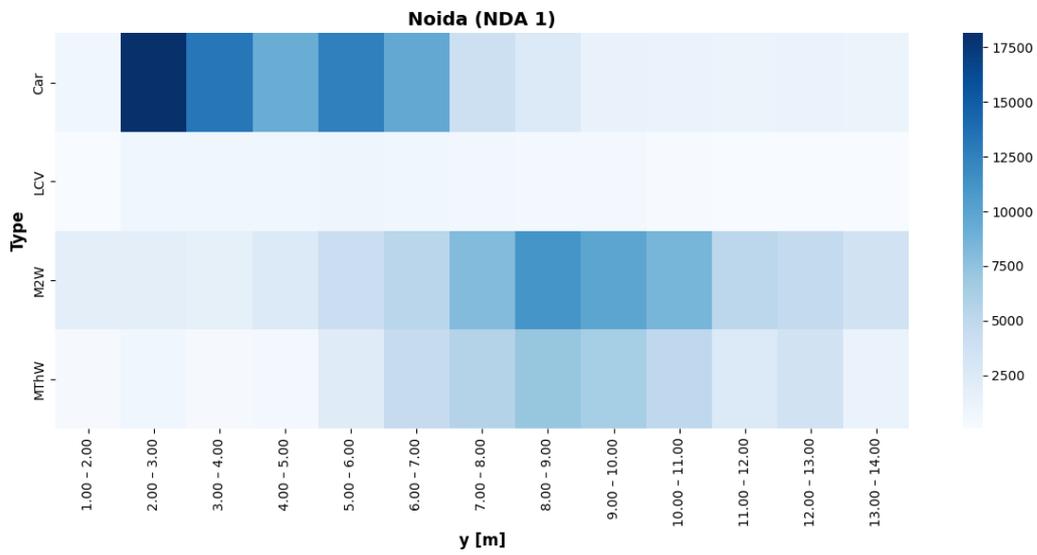
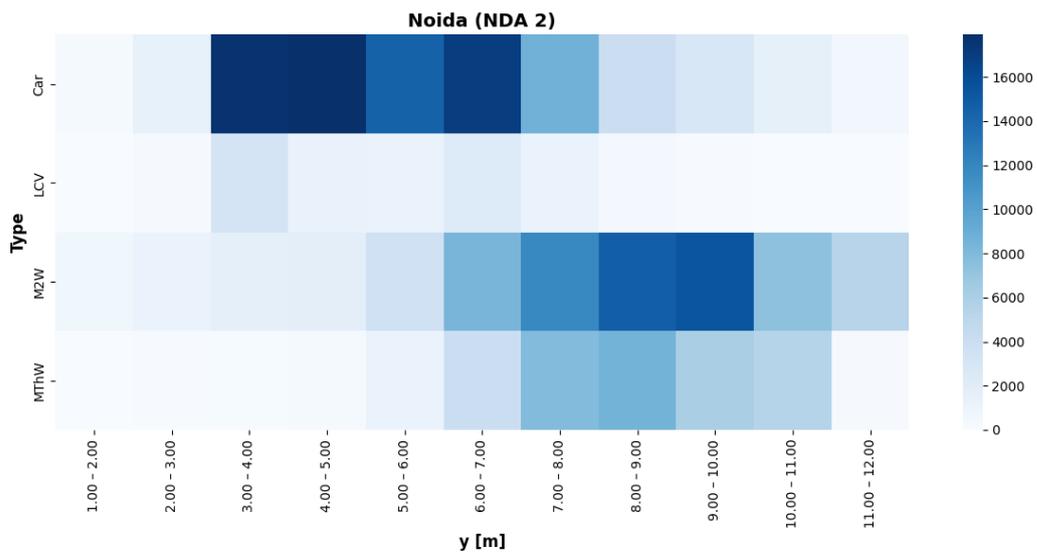

Figure 7 Lane keeping by vehicle types

## 3. Fundamental Diagrams

The following figures visualise the fundamental diagrams (FDs) of traffic flow. The data from all locations are plotted on the FDs to understand the variability in the datasets used in this study. Delhi (DEL_2) displays the expected drop in speed with increasing density, while other sites maintain higher flows at moderate densities. These diagrams illustrate how the datasets encompass a wide range of service levels, from near-free flow to heavy congestion. Such variety makes them suitable for calibrating simulation models across diverse scenarios.

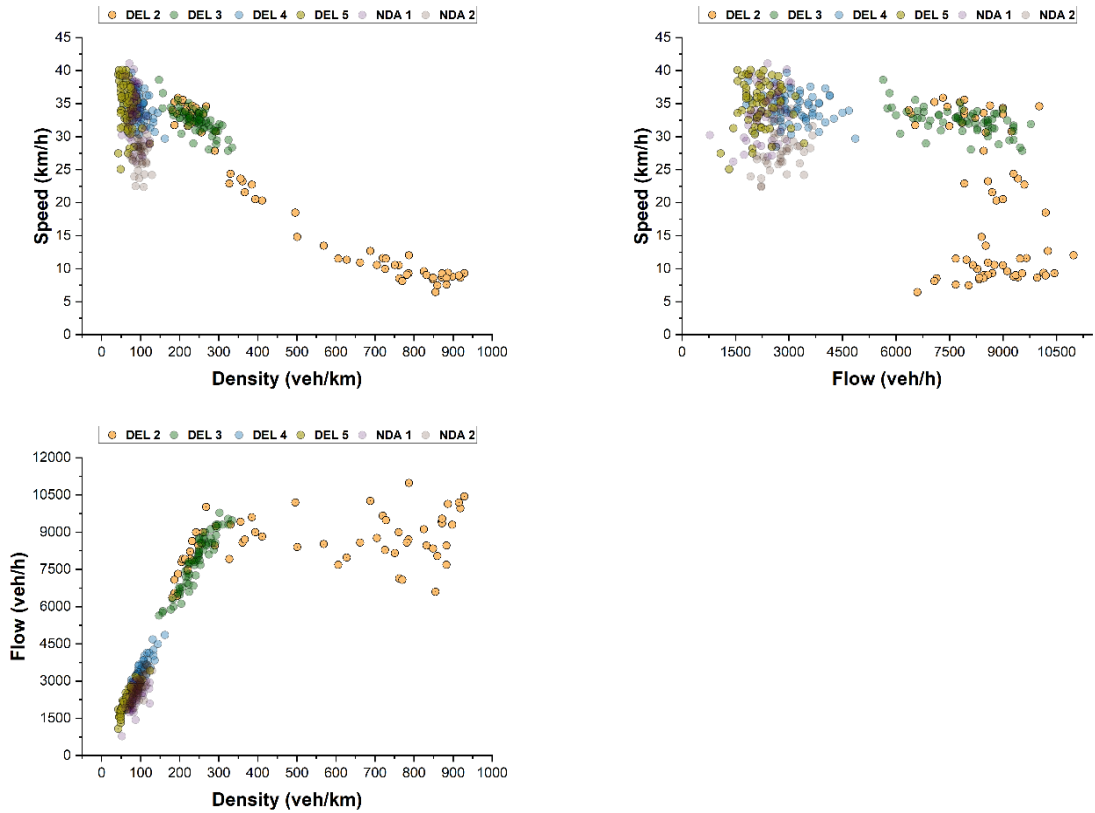

Figure 8 Fundamental diagrams

The rich microscopic vehicle trajectory (MVT) datasets offered in this paper provide high-resolution insights into heterogeneous and area-based urban traffic environments. By releasing them openly, we aim to equip researchers worldwide with data that can advance safety analysis, behaviour modelling, and simulation studies in contexts where lane discipline is weak or disregarded and vehicle interactions are area-based, making the traffic flow more complex than that in the lane-based or homogenous traffic settings. These datasets can provide a foundation for new avenues in traffic studies, such as behavioural modelling, and contribute to the broader impact of making transportation research more transparent, reproducible, and globally relevant.

# CRediT author statement

**Yawar Ali:** Conceptualisation, Methodology, Formal analysis, Investigation, Data Curation, Writing - Original Draft, Writing - Review & Editing, Visualisation. **K. Ramachandra Rao:** Conceptualisation, Investigation, Writing - Review & Editing, Supervision. **Ashish Bhaskar:** Conceptualisation, Investigation, Writing - Review & Editing, Supervision. **Niladri Chatterjee:** Writing - Review & Editing, Supervision.

# Data Availability Statement

All MVT datasets presented in this study can be found and downloaded in their raw form at https://doi.org/10.5281/zenodo.17745347

# Conflict of interest

The authors certify that they have NO affiliation with or involvement in any organisation or entity with any financial interest or non-financial interest in the subject matter or materials discussed that could have appeared to influence the work reported in this manuscript.